\documentclass[lettersize,journal]{IEEEtran}
\usepackage{amsmath}
\usepackage{amsfonts}
\usepackage{algorithmic}
\usepackage{algorithm}
\usepackage{array}
\usepackage[caption=false,font=normalsize,labelfont=sf,textfont=sf]{subfig}
\usepackage{textcomp}
\usepackage{stfloats}
\usepackage{url}
\usepackage{verbatim}
\usepackage{graphicx}
\usepackage{cite}
\usepackage{colortbl}
\usepackage[normalem]{ulem}
\usepackage{multirow}
\usepackage{tabularx}
\usepackage{booktabs}
\usepackage{bbding}
\usepackage[pagebackref=true,breaklinks=true,letterpaper=true,colorlinks,bookmarks=false]{hyperref}
\newlength\savedwidth
\newcommand\whline{\noalign{\global\savedwidth\arrayrulewidth
		\global\arrayrulewidth 1.25pt}%
	\hline
	\noalign{\global\arrayrulewidth\savedwidth}}
\usepackage{ulem}
\usepackage{xcolor}

\begin{document}

\title{Crowd Localization from Gaussian Mixture Scoped Knowledge and Scoped Teacher}

\author{Juncheng~Wang, Junyu~Gao,~\IEEEmembership{Member,~IEEE,} Yuan~Yuan,~\IEEEmembership{Senior Member,~IEEE,} \\and Qi~Wang,~\IEEEmembership{Senior Member,~IEEE}

\thanks{J. Wang is with the School of Software, and with the School of Artificial Intelligence, Optics and Electronics (iOPEN), Northwestern Polytechnical University, Xi'an 710072, P. R. China. E-mail: wangjunchengnwpu@gmail.com.}%
\thanks{J. Gao, Y. Yuan and Q. Wang are with the School of Artificial Intelligence, Optics and Electronics (iOPEN), Northwestern Polytechnical University, Xi'an 710072, P. R. China. E-mails: gjy3035@gmail.com; y.yuan1.ieee@gmail.com; crabwq@gmail.com.} 
\thanks{Q. Wang and J. Gao are the corresponding authors.}
}%

\markboth{IEEE TRANSACTIONS ON IMAGE PROCESSING}%
{Wang \MakeLowercase{\textit{et al.}}: Crowd Localization from Gaussian Mixture Scoped Knowledge and Scoped Teacher}

\maketitle

\begin{abstract}
Crowd localization is to predict each instance head position in crowd scenarios. Since the distance of pedestrians being to the camera are variant, there exists tremendous gaps among scales of instances within an image, which is called the intrinsic scale shift. The core reason of intrinsic scale shift being one of the most essential issues in crowd localization is that it is ubiquitous in crowd scenes and makes scale distribution chaotic.

To this end, the paper concentrates on access to tackle the chaos of the scale distribution incurred by intrinsic scale shift. We propose Gaussian Mixture Scope (GMS) to regularize the chaotic scale distribution. Concretely, the GMS utilizes a Gaussian mixture distribution to adapt to scale distribution and decouples the mixture model into sub-normal distributions to regularize the chaos within the sub-distributions. Then, an alignment is introduced to regularize the chaos among sub-distributions. However, despite that GMS is effective in regularizing the data distribution, it amounts to dislodging the hard samples in training set, which incurs overfitting. We assert that it is blamed on the block of transferring the latent knowledge exploited by GMS from data to model. Therefore, a Scoped Teacher playing a role of bridge in knowledge transform is proposed. What’ s more, the consistency regularization is also introduced to implement knowledge transform. To that effect, the further constraints are deployed on Scoped Teacher to derive feature consistence between teacher and student end.

With proposed GMS and Scoped Teacher implemented on four mainstream datasets of crowd localization, the extensive experiments demonstrate the superiority of our work. Moreover, comparing with existing crowd locators, our work achieves state-of-the-art via F1-measure comprehensively on four datasets.
\end{abstract}

\begin{IEEEkeywords}
Congested Scenes Perception, Crowd Localization, Intrinsic Scale Shift.
\end{IEEEkeywords}

\section{INTRODUCTION}
\IEEEPARstart{C}{rowd} analysis is a popular application to computer vision community and has achieved superb success, especially in crowd counting\cite{song2021choose, yan2019perspective, gao2019pcc, cheng2021decoupled, ling2019indoor, wan2021fine, liu2021PAMI}. Crowd counting is a fundamental task, which estimates the sum counts of pedestrians. The mainstream pipelines produce the predicted counts by directly regressing a scalar\cite{wangQi2022counting} {or integrating the density distribution}\cite{gao2019pcc}. {The above methods cannot yield an accurate location for each instance in crowd scenes, especially in congested regions. Recently, some researchers focus on crowd instance localization} \cite{idrees2018composition, liu2019recurrent, wang2020nwpu, wan2021generalized}, {which aims to locate the center of the head for each 
person. Its instance-level predictions can provide more detailed information than traditional counting algorithms, and it aids some high-level crowd analysis tasks more effectively, such as crowd tracking} \cite{ren2020tracking}, {group detection} \cite{sanford2020group}.

\begin{figure}
    \centering
    \includegraphics[width=0.5\textwidth]{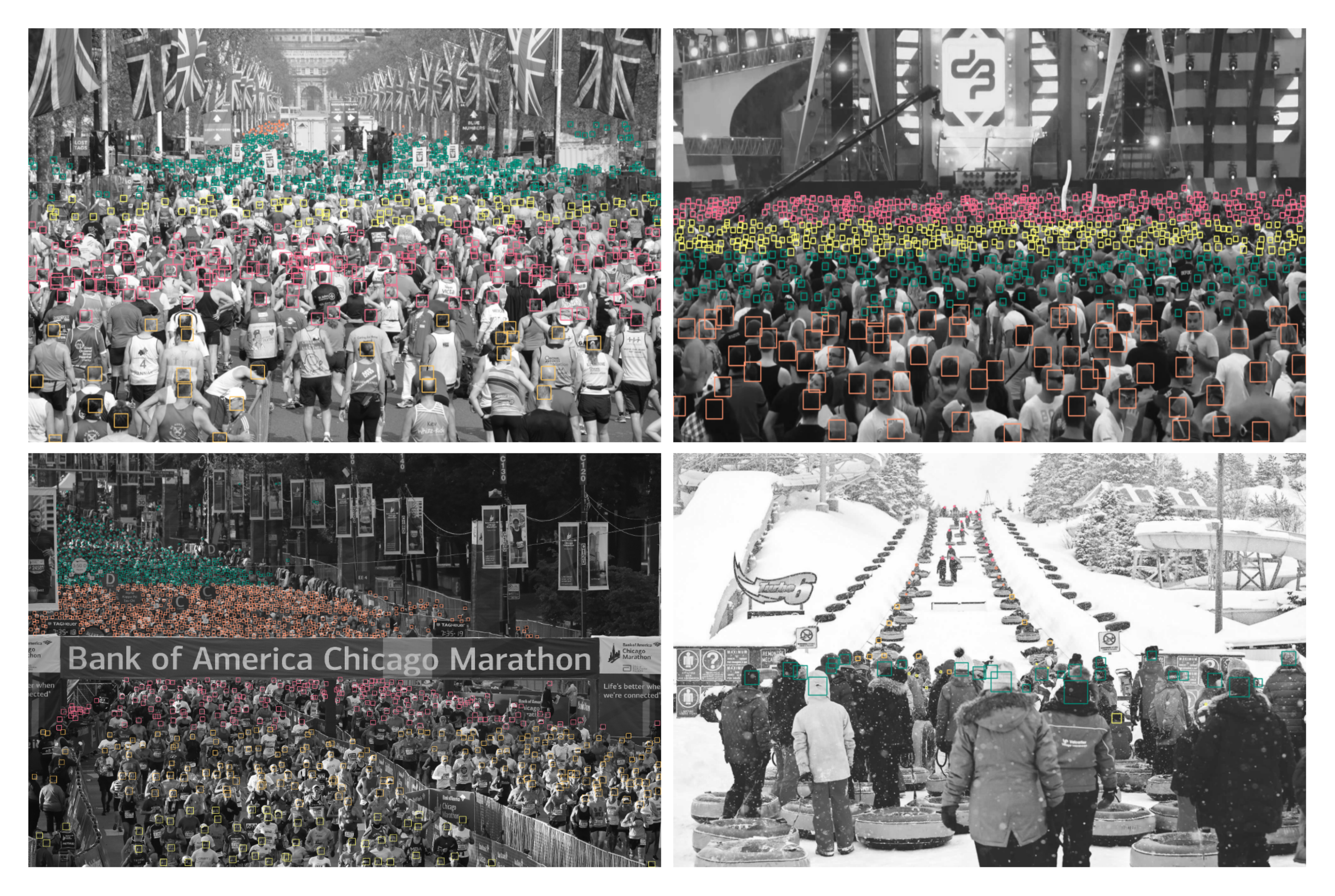}
    \caption{Crowd scenes with intrinsic scale shift. To facilitate visualization, we transfer the image into the gray mode and distinguish boxes with different scales in colors. }
    \label{fig_intro}
\end{figure}

{However, the crowd locator is still a challenging task for its instance-level objective. In crowd scenes}\cite{wang2020nwpu, liu2019recurrent, bai2018finding}{, the instances are represented distinctly in scale within the image, where the representation inconsistence is called intrinsic scale shift.} {The intrinsic scale shift blamed on the different distances for instances from the camera becomes an essential issue in crowd localization for being ubiquitous in crowd scenarios. Fig. \ref{fig_intro} depicts some typical examples for intrinsic scale shift. The boxes with variant scales are annotated in different colors. The intrinsic scale shift makes crowd locator struggle and insufficient to catch instances with variant scales. Precisely, it is arduous for crowd locators to converge on the non-independent identically distributed (\textit{i.i.d.}) data, while the scale distribution of data with intrinsic scale shift can be recognized as the chaotic distribution. Thus, it is imperative to address the intrinsic scale shift in crowd localization.}

{In crowd counting, which is a related but more mature field comparing with crowd localization, the intrinsic scale shift has been attacked with two mainstreams. To begin with model perspective, designing a scale-aware model tackles the intrinsic scale shift in certain.} SAS Net \cite{song2021choose} proposes a fusion strategy among feature maps with different resolutions to aggregate different scales. Despite that the scale-aware models yield a certain promotion, manual proposal to the model architecture is hard to catch certain scale information in the wild. Therefore, the second stream is from data perspective which is to align the intrinsic scale shift. SD Net \cite{ma2021towards} aligns scale shift among orderly divided image patches. However, orderly dividing images ignores the scale variance within the patch. Moreover, the semantic information is distorted in the marginal region of the patches due to patch level dividing. In crowd localization, this semantic distortion of instances 
laying in the marginal region degrades performance. To this end, the crowd locator RAZ Net \cite{liu2019recurrent} leverages a recurrent method to find a region with smallest scales and assign a scale factor to each recurrent layer. Nevertheless, it is challenging to find the smallest scales region without missing other comparing regions.

{This paper aims to tackle the intrinsic scale shift via data regularization and knowledge transform for crowd localization. In data perspective, the intrinsic scale shift incurs the chaos of the scale distribution in crowd scenes.} Thus, we propose a Gaussian Mixture Scope (GMS), which aligns the chaotic scale distribution and constrains the normalization of data. Specifically, a Gaussian mixture distribution is leveraged to adapt to the scale distribution. Through decoupling the feature within the mixture model, the distribution is separated into normal sub-distributions. To this end, the chaos within the scale distribution is mapped to the shift among the sub-distributions.
In the light of the above shift, we utilize the scale alignment among sub-distribution, in which the comparison of probability distributions is geometrized. {Concretely, in adapting the scale distribution via Gaussian mixture, constraining spatial feature as the one of the observation values to probability distribution provides spatial compactness to the sub-distributions}. Therefore, the compactness makes it available to treat sub-distributions as image patches and align the shift via image interpolation.

{Despite that geometrized constraining provides spatial compactness to sub-distributions, the decoupling to the scale distribution incurs certain semantic issues.} Since the decision boundary is adaptive among sub-distributions, the decoupling is to adaptively cut images. With this cutting strategy implemented, the shift alignment via interpolation incurs semantic distorted for distinct scale factors. To this end, we propose a sub-distribution re-aggregation trick. In shift alignment, the images are kept as a whole and fed into crowd locator. The windows are shot from the result according to the corresponding sub-distribution. As a result, there incurs less influence for the undistorted images comparing with distorted ones.

With proposed GMS aligning scale shift and sub-distribution re-aggregation alleviating semantic distortion, the chaos in data distribution is regularized. However, directly implementing GMS in training phase to regular training data dislodges the hard samples. Thus, crowd locator cannot actively \textit{learn} the knowledge, but passively \textit{receive} the knowledge, which incurs overfitting on training set. We assert the GMS regularized data can be treated as exploiting latent knowledge. To further transfer the exploited latent knowledge from regularized data to model, a Scoped Teacher playing a role of bridge in knowledge transform is proposed. The Scoped Teacher introduces a new paradigm comparing with conventional learning from manually annotated ground truth in fully-supervised crowd localization. In training, the GMS regularized images are fed into Scoped Teacher to exploit the latent knowledge, which is hard to be derived from ground truth learning. To transfer the knowledge, a consistency loss is implemented. In this way, the student model gradually learns the Scoped Teacher exploited features and converges better.

In a nutshell, our contributions are four-fold:
\begin{itemize}
\item Propose to tackle the crowd localization from the perspective of scale shift. We provide a novel scale distribution alignment which is to geometrize the issue and to implement it via image interpolation.
\item Present a Gaussian Mixture Scope (GMS) to make scale alignment via scale distribution decoupling and sub-distributions alignment. Moreover, we propose a sub-distribution re-aggregation trick to alleviate boundary semantic distortion in alignment.
\item Design a Scoped Teacher to make latent knowledge transform, which also addresses the overfitting incurred by GMS in direct training. Moreover, the Scoped Teacher is a new paradigm in fully-supervised crowd localization.
\item Quantitative results demonstrate that our proposed work achieves state-of-the-art on four main-stream datasets in crowd localization.
\end{itemize}

\section{RELATED WORKS}
In this section, brief reviews on related works to our method are arrayed. Firstly, since intrinsic scale shift also exists in crowd counting and has been attacked by community, it is of service to review intrinsic scale shift in crowd counting. Secondly, we array the introduction of crowd localization works. At last, to make a distinction with other teacher-student models, we also analyze some representative works which adopt teacher-student architecture.
\subsection{Crowd Counting}
As aforementioned, the counting community attacks intrinsic scale shift in two mainstreams. From the model perspective, a multitude of works \cite{liu2019crowd,liu2018crowd,ma2020learning, cheng2019improving, onoro2016towards, bai2020adaptive, liu2020crowd} deal with intrinsic scale shift via multi-feature fusion. Moreover, some others \cite{shi2019revisiting, yan2019perspective, zhang2015cross, yang2020embedding, liu2019geometric} trace the essence of intrinsic scale shift namely perspective imaging, which utilizes the predicted perspective map as training strategy. Despite that perspective related works achieve certain promotion, we assert the intrinsic scale shift has not been aligned. To this end, Auto Scale \cite{xu2022autoscale} and L2SM \cite{xu2019learn} propose to scale the image patches according to density level. \cite{sajid2020plug, sajid2020zoomcount, babu2017switching} also feed patches with distinct density level into CNN with different receptive fields. \cite{liu2021PAMI} presents the crowd flow to enhance counting performance with location flow. However, the density level cannot represent instance scale. Therefore, SD Net \cite{ma2021towards} introduces instance scale and use it to align scale shift. But SD Net fails to keep semantic information during handle and ignores intrinsic scale shift within the divided patches.

\subsection{Crowd Localization}
Crowd localization aims to locate the precise position of each head shown in the image. The very first idea about the localization must be object detection \cite{redmon2016you, ren2015faster, stewart2016end}. TinyFaces \cite{hu2017finding} utilizes a detection based framework via the analysis to the impacts of scales, context semantic information and image resolution to locate the tiny faces. Following TinyFaces \cite{hu2017finding}, some researchers \cite{bai2018finding, li2019pyramidbox++, li2019dsfd, lin2017focal} make extending work in tackling intrinsic scale shift. However, due to the shortage of detection structure, the detection based methods still perform poorly under extremely congested scenarios. Thus, some researchers begin to utilize regression based crowd locator. RD Net \cite{lian2019density} leverages depth information to generate spatial aware supervision map. But in mainstream datasets of crowd localization, the depth information is unavailable. Thus, \cite{wan2021fine} proposes to utilize fine-grained density map to make crowd localization. BL \cite{wan2021generalized} proposes a location aware loss function to locate the crowds. But it fails to address intrinsic scale shift. \cite{abousamra2021localization, arteta2016counting, gao2020learning, gao2021congested, han2021ldc} utilize instance segmentation to locate crowd. Especially, the instance segmentation locators introduce box annotation in regression. By this way, the instance scale information can be estimated. Thus, our method follows this baseline. What’ s more, there are still some other works concentrating on intrinsic scale shift of crowd localization. Auto Scale \cite{xu2022autoscale} proposes to estimate a density region and learn to zoom it. Similarly, RAZ Net \cite{liu2019recurrent} also proposes a selection strategy to select the density region. These zooming strategies cannot cater to multi-region variance.

\subsection{Teacher-Student Model}
The original proposal of the teacher-student model serves transfer learning. \cite{hinton2015distilling, xie2020self, yang2019training, yuan2021adaptive} utilize teacher-student model in Knowledge Distillation (KD). Actually, our Scoped Teacher model is inspired by KD, in which the teacher model plays a role in bridging data with student model. However, the teacher model in KD tends to utilize a larger teacher model to exploit latent knowledge. However, our Scoped Teacher model shares the same architecture with student model and the images fed into teacher model have been processed by GMS, in which the latent knowledge is not from model representation capacity but from GMS. In Semi-Supervised Learning (SSL), some researchers also introduce teacher model. \cite{sohn2020fixmatch, zhou2021context, yang2021hcdg, ZhuMa2022} introduce teacher model in doing Consistency Regularization. In \cite{araslanov2021self}, they introduce a momentum network to predict pseudo label for unannotated images, which is actually a teacher-student model. The teacher models used in SSL are inclined to predict pseudo labels for unannotated samples which tend to be coarse knowledge. Despite that the proposed teacher model also aims to use Consistency Regularization, our target is to transfer fine-grained knowledge not coarse knowledge which has been learned by student crowd locator with annotation. 

\section{METHODOLOGY}

\begin{figure*}[t]
\centering
\includegraphics[width=1.0\textwidth]{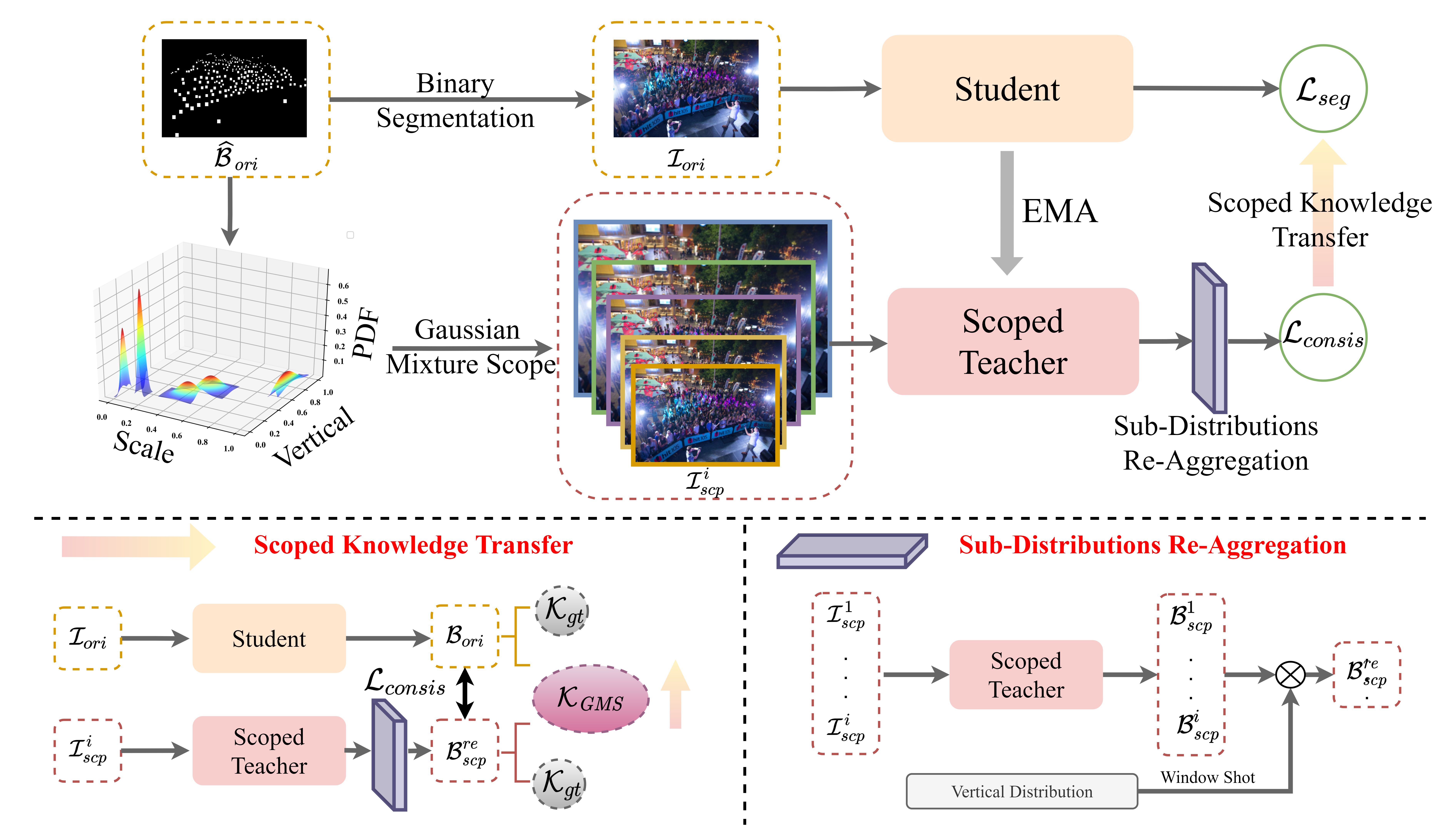} 
\caption{Schematic illustration of our proposed framework. To begin with, we divide pipeline into three branches. The left one denotes the proposed GMS, in which the image is processed before fed into crowd locator. The up stream presents the student end, where the original images are fed. The down stream is the teacher end, in which the images processed by GMS are fed. Finally, a consistency regularization is adopted.} 
\label{fig_pipeline} 
\end{figure*}

\textbf{Overview.} {This paper aims to tackle the intrinsic scale shift in crowd localization. As shown in Fig.} \ref{fig_pipeline}{, we propose a Gaussian Mixture Scope (GMS) to regularize the crowd images to exploit latent scale knowledge. Then, the regularized and original images are fed into proposed Scoped Teacher and student model to make localization prediction.} Then, a sub-distribution re-aggregation is to recompose the scoped prediction. Finally, a consistency loss between predictions by Scoped Teacher and student model is introduced to make knowledge transform. Section \ref{subsecA} reviews the previous Instance Segmentation method, which is our baseline method. Section \ref{subsecB} is for scale alignment process namely proposed GMS and sub-distribution re-aggregation trick. Section \ref{subsecC} describes Scoped Teacher model and knowledge transform. Section \ref{subsecD} gives a summary on our training objective.

\subsection{Instance-Segmentation Crowd Locator}\label{subsecA}
    The popular density map regression method in crowd counting cannot provide precise spatial information. Therefore, some researchers \cite{abousamra2021localization, arteta2016counting} introduce to segment instance-head to make crowd localization. Concretely, they utilize a fixed and global threshold to transfer regressed confidence map activated by a sigmoid function into binary map, which is not robust. Therefore, IIM \cite{gao2020learning} proposes an additional and trainable pixel level threshold map to binarize confidence map. 
    
    Formally, given an image $\mathcal{I}_{ori}\in \mathbb{R}^{3\times H\times W}$, in which the footnote $ori$ represents the original resolution images, a confidence map $\mathcal{F}_{ori}\in \mathbb{R}^{1\times H\times W}$ is predicted, see Eq. \ref{0_1},
    
    \begin{equation}
        \mathbf{0}^{1\times H\times W}\le \mathcal{F}_{ori}\le \mathbf{1}^{1\times H\times W} 
        \label{0_1},
    \end{equation}
    where $ \mathbf{0}$ and $ \mathbf{1}$ denote the tensor filled with $0 / 1$. Additionally, for the fixed threshold works, the segmented binary map $\mathcal{B}_{ori}^{fix}\in \mathbb{R}^{1\times H\times W}$ is obtained through Eq. (\ref{bin}):

\begin{equation}
    \mathcal{B}_{ori}^{fix} \left(h, w\right)=\left\{\begin{array}{lr} 1, & \text { if } \mathcal{F}_{ori}\left(h, w\right) \geq \varepsilon \\
0, & \text { others }
\end{array}\right.
\label{bin},
\end{equation}
where $\varepsilon$ is a fixed threshold and $v, h$ are the pixel coordinates. As for the IIM, the binary map $\mathcal{B}_{ori}^{apt}\in \mathbb{R}^{1\times H\times W}$ is obtained through a trainable threshold map $\mathcal{T}\in \mathbb{R}^{1\times H\times W}$ as Eq. (\ref{IIM}):

\begin{equation}
   \mathcal{B}_{ori}^{apt}\left(h, w\right)=\left\{\begin{array}{lr}
1, & \text { if } \mathcal{F}_{ori}\left(h, w\right) \geq \mathcal{T}\left(h, w\right) \\
0, & \text { others }
\end{array}\right.
\label{IIM}.
\end{equation}

{In Eq. \ref{IIM}, it is obvious that the process is non-differentiable. Thus, \cite{gao2020learning} proposes to relax it to provide a gradient which can be described as} Eq. \ref{BPofIIM}:
\begin{equation}
    \mathcal{T}_{n+1}=\mathcal{T}_n+\alpha \frac{\partial \mathcal{L}_{seg}}{\partial \mathcal{B}},
    \label{BPofIIM}
\end{equation}
{where $\alpha$ is the learning rate and $\mathcal{L}_{seg}$ is formulated as Eq. \ref{lseg}.}

With the adaptative threshold map $\mathcal{T}$, a robust binary map $\mathcal{B}_{ori}^{apt}$ is derived. The training strategy is formulated as Eq. (\ref{lseg}):

\begin{align}
\mathcal{L}_{\text {seg }}= &\frac{1}{H\cdot W} \sum_{h=1}^{H}\sum_{w= 1}^{W}(\left\|\mathcal{F}_{ori}(h,w)-\mathcal{\widehat{B}}(h,w) \right\|^{2}+
\\ &\left\|\mathcal{B}^{apt}_{ori}(h,w) -\mathcal{\widehat{B}}(h,w)\right\|^{1}),
\label{lseg}
\end{align}
where $\mathcal{\widehat{B}}\in \mathbb{R}^{1\times H\times W}$ is the ground-truth binary map of image $\mathcal{I}_{ori}$. By this way, a precise binary map is derived. Therefore, we follow the mentality of IIM as our baseline work. To clarify the paper, we omit the $apt$ in $\mathcal{B}_{ori}^{apt}$ in the following.

\subsection{Gaussian Mixture Scope}\label{subsecB}
In instance segmentation crowd localization, since the locator derives the supervision signal from binary map with implicit scale information, in which the head-areas are annotated as the foreground, the locator is fragile and sensitive to instance scale shift. Moreover, the performance of locator is limited for being hard to catch large and tiny instances simultaneously. {We assert that the issue can be blamed on the chaotically distributed scales.} In the light of deep model trained via Empirical Risk Minimization (ERM), it is arduous for crowd locator to converge on data not satisfing with independent identically distributed conditional assumptions. Summarizing the above analysis, the regularization for the chaotic scale distribution can be the point to tackle the intrinsic scale shift. 

To this end, we propose to decouple the chaotic scale distribution into several regular sub-distributions. Therefore, the chaos within the scale distribution is transferred into the distribution shift among sub-distributions. Additionally, we constrain the spatial feature to be correlated with scale distribution in decoupling. By this way, the sub-distributions are compact in spatial features, and the sub-distributions alignment is available to be implemented via image interpolation. 

{Specifically and formally, given an image $\mathcal{I}_{ori}$ with $N$ pedestrians, in which the footnote $ori$ represents the image in the original resolution, and the corresponding scale distribution $\mathcal{S}_{ori}$ is shown as Eq. (\ref{scaledit}):}

\begin{equation}
    \mathcal{S}_{ori}= \frac{1}{N}\sum_{i=1}^{N}\delta (\alpha _i) 
    \label{scaledit},
\end{equation}
where $\alpha_i$ is the scale for $i^{th}$ instance and $\delta$ denotes one-dimension Dirac function. We utilize a Gaussian mixture distribution to adapt to the $\mathcal{S}_{ori}$ as Eq. (\ref{gmm}):

\begin{equation}
    \mathcal{S}_{ori}\sim \text{Pr}(\alpha ,v|\theta )=\sum_{c=1}^{C}\pi_c \mathcal{N}(\alpha ,v|\mathbf{\theta }) ,
\label{gmm}
\end{equation}
where the mixture distribution is composed of $c$ sub-Gaussian distributions $\mathcal{N}(\cdot)$ with parameters $\theta$ which are mean and variance in Gaussian distribution, and probability $\pi_c$. The $v$ is vertical location to the instance. The mixture model is initialized and updated from the Expectation Maximization, which has an objective function of:

\begin{equation}
    \Theta =\arg \max_{\Theta}\sum_{n}\sum_{c}\lambda_{n,c}\ln{\frac{\pi_c\mathcal{N}(\alpha_n,v_n|\theta _c)}{\lambda _{n,c}}},
    \label{EM}
\end{equation}
in which $\Theta$ is the set of $\{\pi_c, \theta_c|c=1,...,C\}$ and $\lambda$ is defined as Eq. \ref{paraEM}:  
\begin{equation}
    \lambda_{n,c}=\frac{\pi_c\mathcal{N}(\alpha_n,v_n|\theta_c)}{\sum_{c}\pi_c\mathcal{N}(\alpha_n,v_n|\theta_c)}.
    \label{paraEM}
\end{equation}

As aforementioned, to facilitate alignment, the mixture model should be correlated with spatial feature in decoupling. Therefore, we only adopt the vertically spatial feature $v$ to reduce the computational complexity from $\mathcal{O}(n^2) $ to $\mathcal{O}(n) $. As for the horizontal ones, we demonstrate the redundance of it in scale feature representation, see Section \ref{subsubsec ver}. Practically, the fine-grained scale information is unavailable in all existing datasets. Therefore, the annotated box area $\alpha$ is adopted as the observation value to the mixed distribution. 

{After decoupling the mixture distribution, the chaotic scale distribution is decomposed into $c$ normal distributions, which seems to be feasible for model to converge. What’ s more, constraining the vertical features in adaptation and decoupling, the sub-distributions are compact spatially.} Hence, $c$ patches are derived, where each one has a scale distribution of normal sub-distribution in Eq. (\ref{gmm}). To this end, the issue in intrinsic scale shift is to align the scale shift among sub-distributions. Thus, we introduce some prior knowledge, in which an optimal scale $\alpha_0$ is set as the landmark to align the scale shift among sub-distributions. For each patch $p_c$, the aligned one $\widehat{p}_c$ is derived via Eq. (\ref{HHH}) with an interpolation:
\begin{equation}
    \widehat{p_c} =\mathit{Inter}(p_c, \frac{\sum_{i=0}^{N_c} \alpha_i }{\alpha _o\cdot (N_c-1)} ),
\label{HHH}
\end{equation}
where $N_c$ is the count of instance in $p_c$. Note that to avoid computational cost, we make compromise on using average scale of patches. Since the decoupling provides the compactness of scale within the sub-distribution namely the patch, the average scale is adequate to represent the patch.

{Finally, the scale shift is aligned in two levels which are inter-patch and intra-patch.} However, the sub-distributions are still discrete. There are two kinds of normal process, in which one is to directly splice them and make padding on smaller ones, while the other is to keep them being discrete. Nevertheless, in sub-distributions alignment, the patches are interpolated via distinct scale factors, it is unavoidable for the junction region being distorted semantically. Moreover, since the decoupling is adaptive, the decision boundary is uncertain, which incurs that some to be detected instances could be cut off and distorted. To alleviate the issue, we propose a sub-distribution re-aggregation trick.

\begin{figure}[H]
    \centering
    \includegraphics[width=0.5\textwidth]{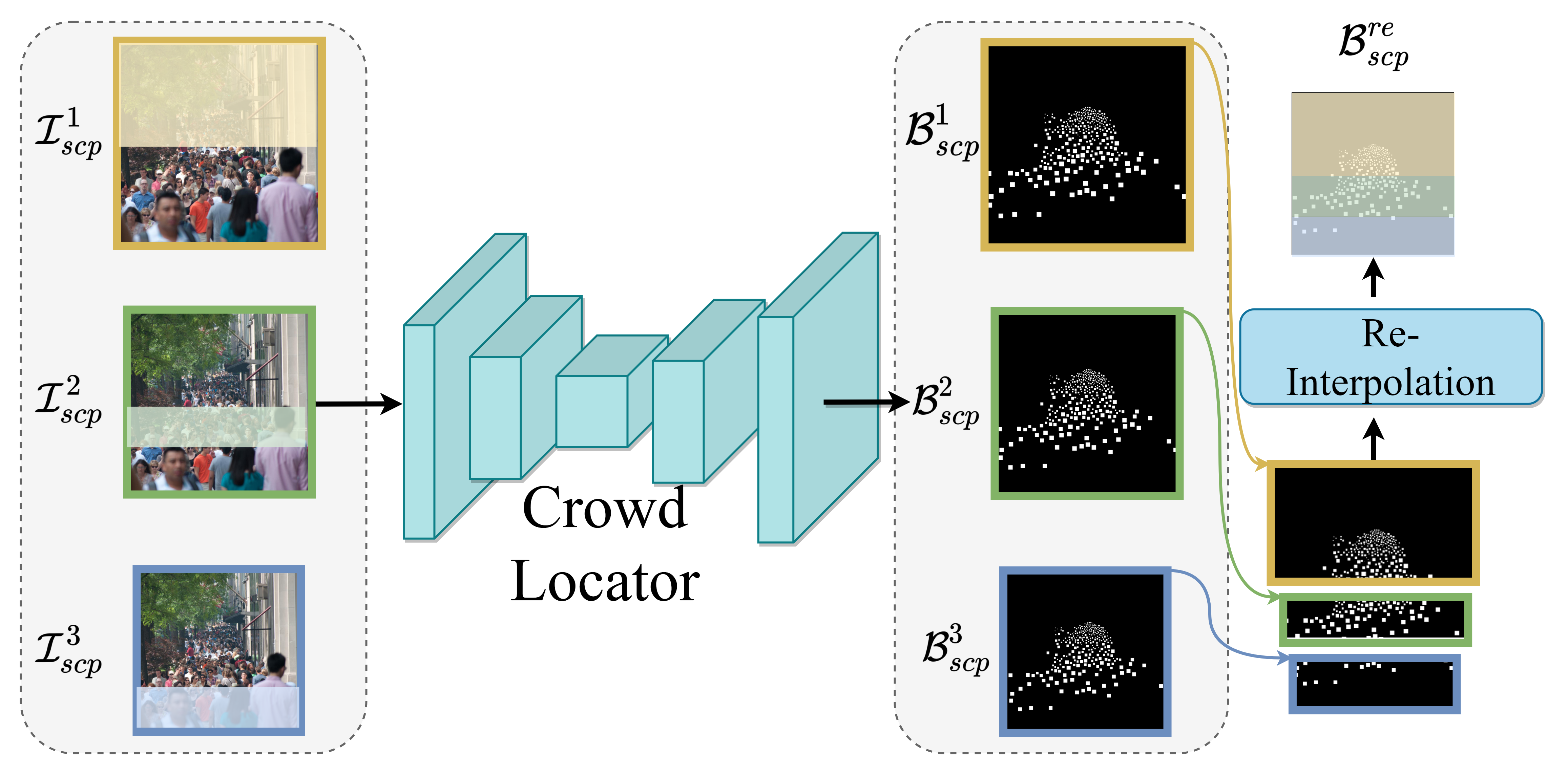}
    \caption{Depiction on the pipeline of sub-distribution re-aggregation. The transparent windows are geometrized sub-distributions. }
    \label{fig_aggregation}
\end{figure}

\textbf{Re-Aggregation for Sub-Distributions.} In aforementioned two processes to discrete sub-distributions, the uncertainty of decision boundary for decoupling incurs the risk for instances being cut off. Given an image or patch with instances cut off, the locator cannot catch the semantic and detect them. Thus, we argue that it is necessary for locator to be fed with whole image. As the Fig. \ref{fig_aggregation} shown, let $\mathcal{I}_{scp}$ be the re-aggregated image. To meet the argument, the Eq. (\ref{eqrelation}) must hold:

\begin{equation}
    \exists \gamma \in \mathbb{R} ,\mathcal{I}_{scp}\equiv Inter(\mathcal{I}_{ori},\gamma ).
    \label{eqrelation}
\end{equation}

Therefore, given $c$ patches with $c$ counts corresponding scale factors $\left \{\alpha _1,\alpha _2,\cdots ,\alpha _c \right \} $ which are from Eq. (\ref{HHH}), we interpolate the $\mathcal{I}_{ori}$ with $\alpha_i$ for $c$ times and derive Eq. (\ref{reagg}):

\begin{equation}
    \left \{ \mathcal{I}_{scp}^i=Inter(\mathcal{I}_{ori}, \alpha_i)\mid i=0,1,\cdots ,c \right \} .
    \label{reagg}
\end{equation}

Then, the $\left \{\mathcal{I}_{scp}^i\mid i=0, 1, \cdots, c\right \}$ are fed into locator which makes it catch correctly semantic information. To obtain the final prediction $\mathcal{B}^{re}_{scp}$ with spatially semantic mapping relation to original image $\mathcal{I}_{ori}$, the predicted $\mathcal{B}^{i}_{scp}$ from $\mathcal{I}_{scp}^i$ are re-interpolated via $\left \{\frac{1}{ \alpha _1},\frac{1}{\alpha _2} ,\cdots ,\frac{1}{\alpha _c}  \right \} $ and the ROI is shot according to $p_c$. The finaly prediction is composed of shot ROIs.

\subsection{Scoped Teacher}\label{subsecC}
In Section \ref{subsecB}, the Gaussian Mixture Scope (GMS) is proposed to regularize the chaotic scale distribution. Given a set of images $\mathcal{I}$ with chaotic scale distribution, the crowd locator represents $\mathcal{I}$ into latent space, in which the partial instances with certain scales are caught and partial backgrounds are mis-embedded. The GMS aligns scale facilitating feature embeddings with variant scale representation to be mapped to the same latent space. {We name the predicted knowledge as $\mathcal{K}_{gt}$ which denotes that they come from ground truth trained model via ERM.} The GMS processes $\mathcal{I}$, in which the outliers not in $\mathcal{K}_{gt}$ are represented to the same latent space and conform $\mathcal{K}_{GMS}$. In training phase, GMS exploits the latent $\mathcal{K}_{GMS}$ to make crowd locator catch the outliers better. However, despite that the implementation of GMS provides additional $\mathcal{K}_{GMS}$ to crowd locator, it is not the active learning, but the passive reception for relationship between $\mathcal{K}_{GMS}$ and locator. As a result, the training phase provides annotation for GMS to exploit  $\mathcal{K}_{GMS}$ and aid locator to perform better, while the annotations are agnostic in testing phase and there is no $\mathcal{K}_{GMS}$ which incurs the locator perform poorly with only representation capacity for $\mathcal{K}_{gt}$. What’ s more in backpropagation phase, GMS dose not have a gradient to compute and its process is non-differentiable. Thus, there should be another better way to deploy GMS and make locator actively learn the $\mathcal{K}_{GMS}$.

To transfer the exploited $\mathcal{K}_{GMS}$, we propose a Scoped Teacher which is a teacher-student framework. Specifically, given an $\mathcal{I}_{ori}$, the student locator has a prediction of $\mathcal{F}_{ori}$ and $\mathcal{B}_{ori}$ which are confidence map and binary map. As for teacher end, the GMS is adopted to regularize the $\mathcal{I}_{ori}$. Then, the processed $\mathcal{I}_{scp}$ is fed into teacher locator to aggregate $\mathcal{K}_{GMS}$ and $\mathcal{K}_{gt}$. The $\mathcal{B}_{scp}^{re}$ is from further proceeding of sub-distribution re-aggregation. Then, a consistency loss is introduced as Eq. (\ref{consis}) to transfer $\mathcal{K}_{gt}$ from Scoped Teacher to student locator.

\begin{align}
\notag \mathcal{L}_{\text {consis}}= &\frac{1}{H\cdot W} \sum_{h=1}^{H}\sum_{w= 1}^{W}(\left\|\mathcal{F}_{ori}(h,w)-\mathcal{B}_{scp}(h,w) \right\|^{2}+
\\ &\left\|\mathcal{B}_{ori}(h,w) -\mathcal{B}_{scp}(h,w)\right\|^{1}).
\label{consis}
\end{align}

In consistency regularization, the Scoped Teacher adopts GMS to exploit $\mathcal{K}_{GMS}$ and restore it in the representation of $\mathcal{B}_{scp}^{re}$. {With Eq. \ref{consis}, the consistency constraint makes the $\mathcal{F}_{ori}$ and $\mathcal{B}_{ori}$ be closer to $\mathcal{B}_{scp}^{re}$. By this way, during training, the back propagation of consistency loss pushes the $\mathcal{K}_{GMS}$ being transferred from Scoped Teacher to student model.}

{Comparing with ground truth supervision, the improvement Scoped Teacher generated is more than GMS exploited knowledge $\mathcal{K}_{GMS}$ transform.} In settings of Scoped Teacher, we leverage a shared architecture with student locator. Empirically, to utilize a larger model in teacher end designing could make teacher with stronger representation capacity guide weaker student training. However, our Scoped Teacher shares the same architecture to student model, which means the outputs between student and teacher ends are with more consistent. Therefore, the guidance of Scoped Teacher to student is feasible to implement consistency regularization. 

Finally, in parameters updating, the student crowd locator is trained via gradient descend. To aggregate the knowledge and stable knowledge transform, the teacher parameters $\vartheta_t$ is updated via Exponential Moving Average (EMA) with student parameters $\vartheta _s$ as Eq (\ref{EMA}):
\begin{equation}
    \vartheta_t\gets m\vartheta_t + (1-m)\vartheta_s,
\label{EMA}
\end{equation}where $m$ denotes the EMA decay coefficient to control the updating rate.

\subsection{Objective}\label{subsecD}
\textbf{Instance Segmentation Loss.} The instance segmentation loss is a L2 loss for confidence map regularization and a L1 loss for binary map regularization as Eq. (\ref{lseg}).

\textbf{Consistency Regularization Loss.} Since the gradient is detached in threshold learner, to optimize the threshold learning, L1 loss is used for binary map regularization between teacher and student end as Eq. (\ref{consis}).

\textbf{Total Loss.} During training, the student model is jointly trained in an end-to-end manner. The whole parameters are updated by integrating all mentioned loss functions:

\begin{equation}
    \mathcal{L}_{total}=\mathcal{L}_{seg} + \mathcal{L}_{consis}.
\end{equation}
The teacher model is optimized as Eq. (\ref{EMA}).

\subsection{Inference}
{At the testing or inference phase, the original image $\mathcal{I}_{ori}$ is fed into student model, which can be described as Eq. \ref{inference}:}
\begin{equation}
    \mathcal{B}_{res}=f(\mathcal{I}_{ori};\vartheta_{t}).
    \label{inference}
\end{equation}

{Therefore, our proposed method would not incur any extra costs in inference.}

\section{Experiment}\label{sec_exp}
\subsection{Datasets}
\begin{itemize}
\item[1] \textbf{Shanghai Tech Part A (SHHA):} SHHA \cite{zhang2016single} contains 482 images where 270 are available for training, 30 are for validation and others are for test. There are 241, 677 instances annotated in total.
\item[2] \textbf{Shanghai Tech Part B (SHHB):} SHHB \cite{zhang2016single} consists of 716 images where 360 are prepared for training, 40 are for validation and others are for test. SHHB has 88, 488 annotated instances.
\item[3] \textbf{NWPU-Crowd (NWPU):} NWPU \cite{wang2020nwpu} is the largest dataset in crowd analysis community so far, in which there contains 5,109 images with 3,109 of them for training, 500 for validation and 1,500 for test. 
\item[4] \textbf{UCF-QNRF (QNRF):} QNRF \cite{idrees2018composition} is a dataset with extremely congested scenarios, where it is composed of 1,535 images and 961 of them are for training, 240 are for validation and others are for test.
\end{itemize}

\subsection{Implementation Details and Metrics}
{In the training phase, backbone networks of VGG-16} \cite{simonyan2014very} {and HR-Net} \cite{wang2020deep}{, a batch size of 6, an optimizer of Adam} \cite{kingma2015adam} {with learning rates 1e-5 for backbone and 1e-6 for threshold encoder, a learning rate decay of 0.99 for every epoch are adopted, an interpolation mode of Bilinear}. In the testing phase, the tested images are fed into student locator in original scale and the model with best performance on validation set is picked for testing. Moreover, our experiments are applied on two NVIDIA RTX 3090 with a total memory of 48 GB.

Following \cite{gao2020learning, wang2020nwpu}, the Precision (Pre.) , Recall (Rec.) and F1-measure (F1-m) are adopted for localization metrics as Eq. (\ref{F1}),

\begin{align}
    \notag Pre. &= \frac{TP}{TP+FP},\\
    \notag Rec. &= \frac{TP}{TP+FN},\\
    F1 \text{-}m &= \frac{2\cdot Pre\cdot Rec}{Pre+Rec},
\label{F1}
\end{align} where F1-m is the core norm and $TP, TN, FP, FN$ denote True Positive, True Negative, False Positive and False Netgative. The MAE, MSE and NAE are adopted for counting metrics as Eq. (\ref{MAE}),

\begin{align}
\notag MAE&=\frac{1}{N}  \sum_{i=1}^{N}\left \| z_i-\widehat{z}_i \right \|^1,\\ \notag MSE&=\sqrt{\frac{1}{N}\sum_{i=1}^{N}\left \| z_i-\widehat{z}_i  \right \|^2},\\
NAE&=\frac{1}{N} \sum_{i=1}^{N}\frac{\left \| z_i-\widehat{z}_i  \right \|^1 }{z_i}.
\label{MAE}
\end{align}

\subsection{Analysis on Our Method}
\subsubsection{Ablation study}\label{ablation}
In this section, our method is decomposed into components to exploit each contribution. Individually, we only implement our proposed GMS in the inference phase to explore whether it is effective to align the intrinsic scale shift. Then, the same teacher model called Plain Teacher in Tab. \ref{tab AA} without GMS is introduced as demonstration in the training phase. The same consistency regularization loss is also utilized. Finally, our whole system is deployed.

\begin{table}[h]
\centering

\caption{Ablation study tested on SHHA-val.}
\setlength{\tabcolsep}{3mm}
\renewcommand\arraystretch{1.2}
\begin{tabular}{c|c|c} 
\whline
\multirow{2}{*}{Method} & Localization                       & Counting                       \\ 
\cline{2-3}
                        & \textbf{F1-m}/Pre./Rec. (\%)       & MAE/MSE                        \\ 
\whline
Baseline                & 67.0 /71.0 /63.4                   & 119.5/ 242.1                   \\ 
\hline
GMS Inference           & 69.2/74.1/ 64.8     & 85.1 /164.8                    \\ 
\hline
Plain Teacher           & 69.1 /\textbf{75.9}/ 63.3                   & 119.5 /260.4                   \\ 
\hline
Whole Method            & \textbf{71.4}/73.6/ \textbf{69.3~} & \textbf{81.7 }/\textbf{147.1}  \\
\whline
\end{tabular}
\label{tab AA}
\end{table}

In Tab. \ref{tab AA}, baseline model is to leverage a pixel level threshold map to binarize confidence map into binary map. GMS Inference directly implements GMS in the inference phase and experiment shows that GMS promotes the localization performance by 2.2$\%$ and the counting performance by 34.4 on MAE. Therefore, GMS is indeed effective to align the intrinsic scale shift. Then, the Plain Teacher model with ensemble learning also has promotion which can be attributed to the knowledge aggregation. In the whole system implementation, our proposed method makes large promotion on both localization and counting.

\subsubsection{Effect on knowledge transform} The proposed GMS is an off-line regularization strategy. In Section \ref{ablation}, we demonstrate that directly implementing GMS in inference promotes the performance. However, the implementation of GMS requires ground truth of samples and incurs additional computing overhead. Thus, we propose Scoped Teacher to transfer the GMS exploited knowledge. To see if the knowledge GMS exploited transferred to student locator, we implement experiments on Tab. \ref{KT}. 

\begin{table}[h]
\centering
\caption{Demonstration on Knowledge Transform.}

\begin{tabular}{c|c|c} 
\whline
\multirow{2}{*}{Method} & Localization                 & Counting             \\ 
\cline{2-3}
                        & \textbf{F1-m}~/ Pre. / Rec.  & \textbf{MAE }/ MSE~  \\ 
\whline
Base                    & 67.0 / 71.0 / 63.4           & 119.5 / 242.1        \\
Base + GMS              & 69.2 / 74.1/ 64.8            & 85.1 /164.8          \\
Improvement             & +2.2 / +3.1 / +1.4           & +34.4 / 77.3         \\ 
\hline
Scoped                  & 71.3 / 74.3 / 68.6\textbf{~} & 86.4 / 163.8         \\
Scoped + GMS            & 71.6 / 74.9 / 68.4           & 79.5 / 152.5         \\
Improvement             & +0.3 / +0.6 / -0.2           & +6.9 / +11.3         \\
\whline
\end{tabular}
\label{KT}
\end{table}

In Tab. \ref{KT}, the \textit{Base} denotes the baseline crowd locator, \textit{GMS} denotes to implement GMS aligning testing data online. \textit{Scoped} means the model is trained via our whole Scoped Teacher. Specifically, we implement GMS at testing time on locators knowledge transferred and without knowledge transferred. The results show that GMS regularization is effective in improving localization and counting performance to baseline model, in which there are marginal improvement obtained. However, there is slight influence when GMS is implemented on Scoped Teacher transferred locator. It is demonstrated that the Scoped Teacher transferred locator indeed learn the GMS exploited knowledge. The additionally deployed GMS is useless in knowledge extraction.
\subsubsection{Choice of prior optimal scale} In GMS implementation, an optimal scale is introduced to compute an optimal interpolation factor for each sub-distribution. Intuitively, the optimal scale should be as large as possible. However, a considerable scale incurring large resolution is computational in convolution process. Moreover, image interpolation with a huge factor incurs serious non-semantic distortion. Therefore, some scales are selected to draw a finally optimal scale.

\begin{table}
\centering
\caption{Choice on different optimal scales.}
\renewcommand\arraystretch{1.2}
\setlength{\tabcolsep}{3mm}
\begin{tabular}{c|c|c} 
\whline
\multirow{2}{*}{Optimal Scale} & Localization                  & Counting               \\ 
\cline{2-3}
                               & \textbf{F1-m} /Pre./ Rec.(\%) & MAE /MSE               \\ 
\whline
100                            & 67.9/ \textbf{75.2}/61.9     & 111.9 /227.6          \\ 
\hline
250                            & \textbf{69.2}/74.1/ 64.8     & 85.1 /164.8            \\ 
\hline
500                            & 68.9 /72.7 /65.6            & \textbf{76.9}/156.4   \\ 
\hline
1,000                           & 67.6/ 68.9 /66.3              & 80.3/\textbf{148.3~}  \\ 
\hline
5,000                           & 62.1/ 57.4/ \textbf{67.7~}    & 163.3/ 211.8          \\
\whline
\end{tabular}
\label{optimal}
\end{table}

With the optimal scale being larger in Tab. \ref{optimal}, the performance is not positively and correlatedly varying. We find a moderate scale is the best for performance promotion. For the smaller scale, the tiny instances are under zoomed. The locator tends to pay more attention on easy scales but ignore tiny instances, which is reflected on high precision, low recall and terrible counting performance. For the huge scales, we argue that it yields extreme distortion, which is shown on over-estimation. Thus, the Precision is low but Recall is high under huge scales. At last, the 250 and 500 are comparative. An optimal scale of 250 is finally chosen. This is because larger scale incurs higher computational complexity. 

\subsubsection{Comparison on three sub-distribution processing strategies} In this section, we compare three kinds of interpolation methods during inference phase to demonstrate the effect of proposed Re-Aggregation. Firstly, the image is divided into patches as GMS decoupled. Then, the patches are fed into crowd locator successively and the results have been arrayed as Patch Divide in Tab. \ref{patches}. Secondly, based on Patch Divide, the patches are spliced into a hierarchically arrayed image, whose results have been arrayed as Patch Whole. Finally, our proposed Re-Aggregation is shown. 
\begin{table}[h]
\centering
\caption{Comparison among three kinds of sub-distribution processes.}
\renewcommand\arraystretch{1.2}
\setlength{\tabcolsep}{3mm}
\begin{tabular}{c|c|c}
\whline
\multirow{2}{*}{Method} & Localization                     & Counting                       \\ 
\cline{2-3}
                        & \textbf{F1-m}/Pre./Rec. (\%)     & MAE/MSE                        \\ 
\whline
Baseline                & 67.0/71.0/63.4                   & 119.5/242.1                    \\ 
\hline
Patch Divide            & 63.5/64.4/62.7                   & 87.0/130.4                     \\ 
\hline
Patch Whole             & 68.2/68.8/\textbf{\textbf{67.7}} & \textbf{76.3}/\textbf{118.27}  \\ 
\hline
Re-Aggregation      & \textbf{69.2}/\textbf{74.1}/64.8 & 85.1/164.8 \\
\whline
\end{tabular}
\label{patches}
\end{table}
\\
According to results, our Re-Aggregation performs best on F1-m but fails on counting performance. Therefore, we analyze the binary map from methods. We notice that in the marginal regions, the instances semantic information is distorted. To this end, the heads laying on the boundary line are divided into two parts. Thus, an additional prediction is generated. The counting results are higher which is closer than ground truth count. Moreover, for the imbalanced dividing, the patch with bigger part of heads cannot represent true position, which incurs the corresponding prediction to be recognized as False-Positive. Therefore, the Precision of Patch Divide and Patch Whole is even lower than Baseline model. In summary, our Re-Aggregation indeed alleviates the semantic distortion in the marginal region.
\begin{table*}[t]
\centering
\caption{The leaderboard of NWPU-Crowd Localization (test set).}
\renewcommand\arraystretch{1.5}
\setlength{\tabcolsep}{5mm}
\begin{tabular}{c|c|c|c|c|c} 
\whline
\multirow{2}{*}{Method} & \multirow{2}{*}{Backbone} & \multicolumn{2}{c|}{Overall Performance}                                    & \multicolumn{2}{c}{Scale Level}                                                                 \\ 
\cline{3-6}
                                                                     &                           & \textbf{F1-m}/Pre/Rec(\%)               & MAE/MSE/NAE                       & Avg.          & A0\textasciitilde{}A5                                                           \\ 
\whline
Tiny Faces                                                           & ResNet-101                & 56.7/52.9/61.1                          & 272.4/764.9/0.750                 & 59.8          & 4.2/22.6/59.1/\textbf{90.0}/\textbf{93.1}/\uline{89.6}                         \\ 
\hline
RAZ\_Loc                                                             & VGG-16                    & 59.8/66.6/54.3                          & 151.5/634.7/0.305                 & 42.4          & 5.1/28.2/52.0/79.7/64.3/25.1                                                    \\ 
\hline
VGG+GPR                                                              & VGG-16                    & 52.5/55.8/49.6                          & 127.3/439.9/0.410                 & 37.4          & 3.1/27.2/49.1/68.7/49.8/26.3                                                    \\ 
\hline
IIM                                                                  & VGG-16                    & 73.2/77.9/69.2                          & 96.1/414.4/0.235                  & 58.7          & 10.1/44.1/70.7/82.4/83.0/61.4                                                   \\ 
\hline
TopoCount                                                            & VGG-16                    & 69.2/68.3/70.1                          & 107.8/438.5/-                     & \uline{63.3}  & 5.7/39.1/72.2/85.7/87.3/\textbf{89.7}                                                    \\ 
\hline
AutoScale                                                            & VGG-16                    & 62.0/67.4/57.4                          & 123.9/515.5/0.304                 & 48.4          & 4.1/29.7/57.2/76.1/78.7/44.6                                                    \\ 
\hline
Ours                                                                 & VGG-16                    & 74.3/80.8/68.7                          & 102.9/446.8/0.245                 & 60.3          & 10.7/42.6/69.8/83.3/86.2/69.0                                                   \\ 
\whline
Crowd-SDNet                                                          & ResNet-50                 & 63.7/65.1/62.4                          & -/-/-                             & 55.1          & 7.3/43.7/62.4/75.7/71.2/70.2                                                    \\ 
\hline
FIDTM                                                                & HR-Net                    & 75.5/79.8/71.7                          & 86.0/\textbf{312.5}/0.277         & 47.5          & \textbf{22.8}/\textbf{66.8}/\uline{76.0}/72.0/37.4/10.3                        \\ 
\hline
IIM                                                                  & HR-Net                    & 76.2/\uline{81.3}/71.7                  & 87.1/406.2/\textbf{0.152}         & 61.3          & 12.0/46.0/73.2/85.5/86.7/64.3                                                   \\ 
\hline
DCST                                                                 & Swin-ViT                    & \uline{77.5}/\textbf{82.2}/\uline{73.4} & \textbf{84.2}/374.6/\uline{0.153} & 60.9          & 14.5/51.0/75.3/85.0/81.7/57.8            \\ 
\hline
Ours                                                                 & HR-Net                    & \textbf{78.1}/79.8/\textbf{76.5}        & \uline{84.7}/\uline{361.5}/0.232  & \textbf{66.7} & \uline{17.1}/\uline{54.1}/\textbf{78.0}/\uline{88.0}/\uline{90.6}/72.3  \\
\whline
\end{tabular}
\label{TabNWPPUU}
\end{table*}


\begin{figure}[htbp]
    \centering\includegraphics[width=0.5\textwidth]{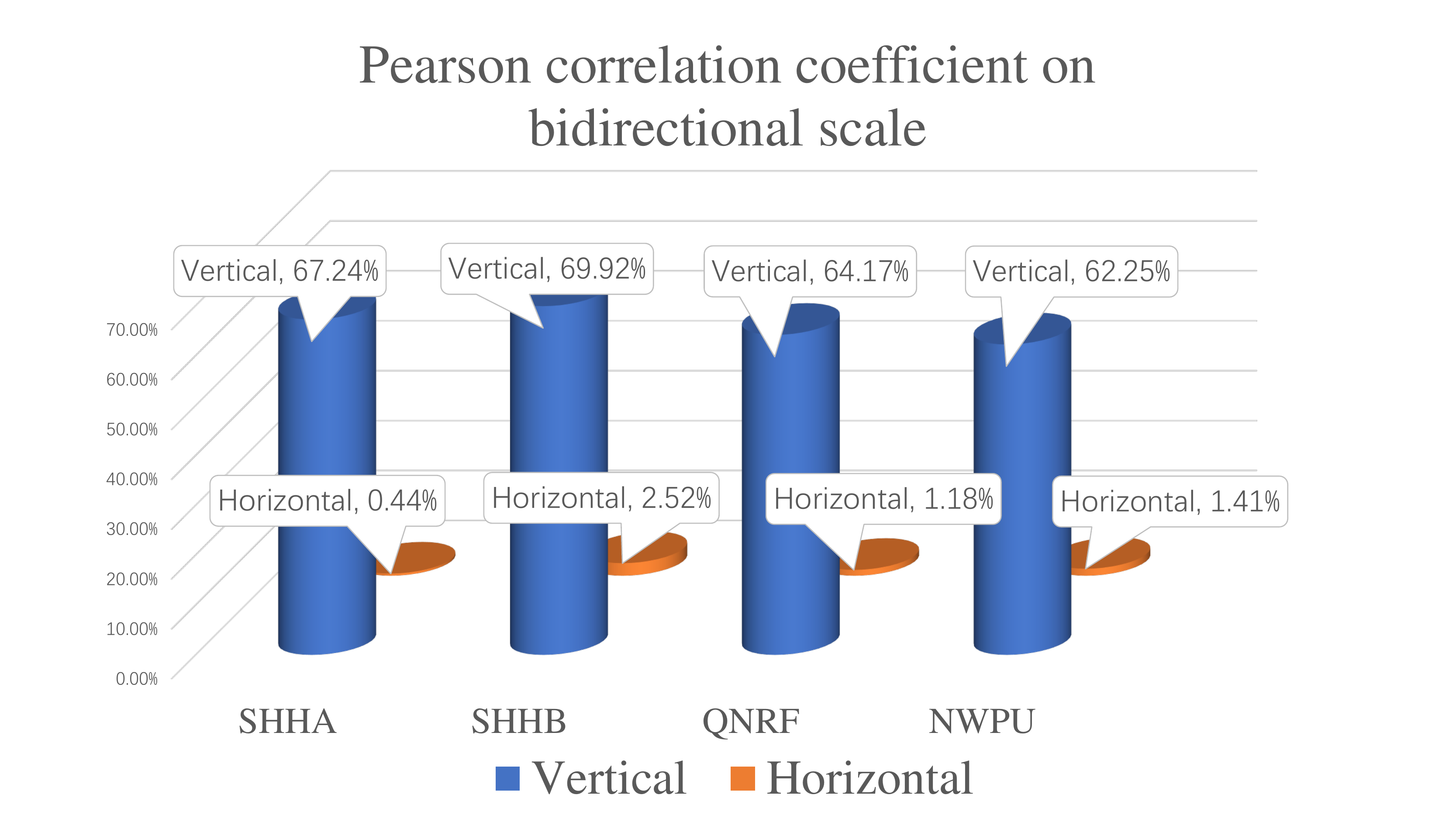}
    \caption{Pearson correlation coefficient on scale with two directions on four adopted datasets.}
    \label{pearson}
\end{figure}

\subsubsection{Why did only vertical features work}\label{subsubsec ver}
In crowd scenes, the scale distribution is inclined to be correlated with spatial distribution. This is caused by imagining process, the adjacent instances in physical space have similar scale in image. In our setting, the adaptation in scale distribution further introduces spatial feature to facilitate scale alignment via image interpolation. However, introducing spatial feature from two directions namely vertical and horizontal ones incurs a computational complexity of $\mathcal{O}(n_v\cdot n_h)$, where $n_v$ and $n_h$ denote the number of sub-distributions in vertical and horizontal direction. From the point of saving training cost, we analyze how vital for some direction in scale representation. To this end, we introduce Pearson correlation coefficient to measure how correlated between scale with the two spatial features.

In Tab. \ref{pearson}, the correlation coefficients between scale with vertical feature and horizontal feature show that the horizontal feature is almost independent with scale. In adaptation, we aim to utilize spatial feature to represent scale. Thus, the horizontal feature is slight in our objective.

\subsection{Comparison with State-of-the-Art Methods}
In this section, four chosen datasets are grouped into three parts. NWPU and JHU are comprehensive dataset where the congested and sparse scenarios are all included. QNRF and SHHA are congested datasets, while SHHB is the sparse dataset.
\subsubsection{Comparison with SOTA methods on comprehensive datasets}
In this section, we compare our proposed method with SOTA methods on NWPU-Crowd. 

Tab. \ref{TabNWPPUU} arrays the comprehensive results on Localization and Counting on NWPU. In Tab. \ref{TabNWPPUU}, the chosen methods are divided as their used backbone network for a fair comparison. The Scale Level norm is Recall value. A0\textasciitilde{}A5 denotes the instance-scale is in [$10^0, 10^1$], ($10^1, 10^2$], ($10^2, 10^3$], ($10^4, 10^5$] and ($10^5$, +$\infty$). The bold text denotes the first place and the underlined text denotes the second place.The compared methods are TinyFaces\cite{bai2018finding}, RAZ$\_$Loc\cite{liu2019recurrent}, VGG+GPR\cite{gao2019domain, gao2019c}, IIM\cite{gao2020learning}, TopoCount\cite{abousamra2021localization}, AutoScale\cite{xu2022autoscale}, Crowd-SDNet\cite{wang2021self}, DCST\cite{gao2021congested}. Additionally, TinyFaces and Crowd-SDNet utilize \cite{he2016deep} as backbone network. With comparing primary norms (F1-m and MAE), our proposed method achieves the \textbf{first place} on Localization performance (a F1-m of 78.1$\%$). Furthermore, the Recall value on different scales is also arrayed. Tab. \ref{TabNWPPUU} shows that our work is the first or second place on almost all scales. 

\begin{figure*}[t]
\centering
\includegraphics[width=1.0\textwidth]{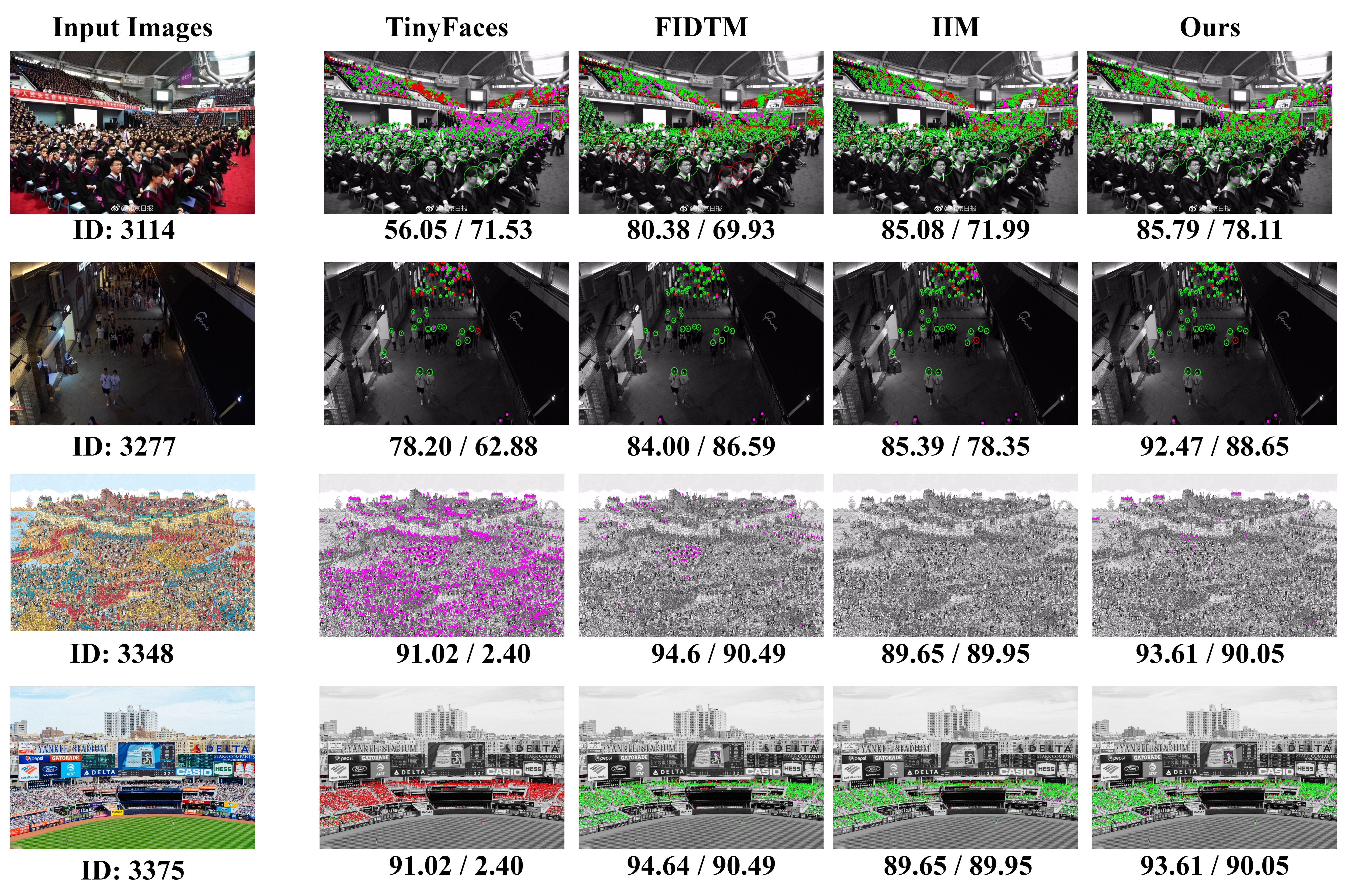} 
\caption{
Qualitative results on the NWPU-Crowd validation set. The predicted TP, FN and FP are respectively denoted as green, red and magenta. The results on the top of each sample have a template of $Pre. / Rec.$} 
\label{figsum}
\end{figure*}

What’ s more, we intuitively depict localization results. Following \cite{wang2020nwpu}, we pick three representative methods to compare with our methods. Concretely, TinyFaces \cite{bai2018finding} is the object detection crowd locator. FIDTM \cite{liang2021focal} is the density regression crowd locator. IIM \cite{gao2020learning} is the instance segmentation crowd locator. Fig. \ref{figsum} illustrates four groups of typical samples, in which the $3114^{th}$ is the low resolution scene, $3277^{th}$ is the sparse scene, $3348^{th}$ is the negative scene and $3375^{th}$ is the extremely congested scene. Firstly, for the low resolution scene namely the region in the top of $3114^{th}$, our Secoped Teacher performs better than others, duo to the zoom strategy to the tiny scales. Secondly, the sparse scenes like $3277^{th}$ tend to suffer more serious scale shift empirically. Thus, we surpass the others in an untrivial margin. Thirdly, our scale alignment dose not break the robustness under the negative scenes, \textit{i.e.}, $3348^{th}$. As last, in extremely congested scenes, the density regression based FIDTM performs best. In the instance segmentation based locator, the congested scenarios incur tremendous overlapping, so the performance is relatively poor.

\subsubsection{Comparison with SOTA methods on Congested Datasets}
In this section, we compare our proposed method with other five state-of-the-art crowd locators in two congested datasets (QNRF and SHHA). The compared locators are TinyFaces, RAZ Loc, LSC-CNN, IIM and DCST. Specifically, TinyFaces is trained via official project with default parameters. RAZ Loc is adopted from \cite{wang2020nwpu}. LSC-CNN and IIM also come from official implementation. The performance of DCST is from arxiv preprinted paper. The performance (Localization: F1-m, Precision, Recall; Counting: MAE and MSE) are arrayed in Tab. \ref{qnrf}. 
In SHHA, our proposed method achieves first place on F1-m and second place on MAE. Comparing with instance segmentation crowd locators IIM and DCST, we outperform them (76.0$\%$ vs. 73.9$\%$ and 74.5$\%$) only with VGG-16 being backbone network. In QNRF, our work achieves first place on localization and counting. Significantly, the VGG-16 version of our work surpasses Swin-Transformer \cite{liu2021swin} based DCST (72.6$\%$ vs. 72.4$\%$).

\begin{table*}[t]
\centering
\caption{Comparison with SOTA methods on congested datasets.}
\renewcommand\arraystretch{1.5}
\setlength{\tabcolsep}{5mm}
\begin{tabular}{c|c|c|c|c|c} 
\whline
\multirow{2}{*}{Method} & \multirow{2}{*}{Backbone} & \multicolumn{2}{c|}{QNRF}                                    & \multicolumn{2}{c}{SHHA}                                              \\ 
\cline{3-6}
                        &                           & \textbf{F1-m}/Pre./Rec. (\%) & MAE/MSE                       & \textbf{\textbf{F1-m}}/Pre./Rec. (\%) & MAE/MSE                       \\ 
\whline
TinyFaces               & ResNet-101                & 49.4/36.3/\textbf{77.3~}     & 336.8/741.6                   & 57.3/43.1/\textbf{85.5~}              & 237.8/422.8                   \\ 
\hline
RAZ\_Loc                & VGG-16                    & 53.3/59.4/48.3               & \underline{118.0}/\underline{198.0}                   & 69.2/61.3/\underline{79.5}                        & 71.6/\underline{120.1}                    \\ 
\hline
LSC CNN                 & VGG-16                    & 58.2/58.6/57.7               & 120.5/218.3                   & 68.0/79.6/66.5                        & \textbf{66.4}/\textbf{117.0}  \\ 
\hline
IIM                     & VGG-16                    & 68.8/\underline{78.2}/61.5               & 160.6/290.0                   & 72.5/72.6/72.5                        & 83.6/164.2                    \\ 
\hline
Ours                    & VGG-16                    & \underline{72.6}/77.0/68.7               & 137.6/263.2                   & \underline{76.0}/76.4/75.5                        & 71.8/128.1                    \\ 
\whline
IIM                     & HR-Net                    & 72.0/\textbf{79.3}/65.9      & 142.6/261.1                   & 73.9/\underline{79.8}/68.7                        & 69.3/138.7                    \\ 
\hline
DCST                    & Swin-ViT                    & 72.4/77.1/68.2               & 127.2/234.3                   & 74.5/77.2/72.1                        & 78.4/153.2                    \\ 
\hline
Ours                    & HR-Net                    & \textbf{75.5}/77.9/\underline{73.4}      & \textbf{104.4}/\textbf{197.4} & \textbf{78.1}/\textbf{81.7}/74.9      & \underline{68.8}/138.6                    \\
\whline
\end{tabular}
\label{qnrf}
\end{table*}
\subsubsection{Comparison with SOTA methods on Sparse Dataset SHHB}
In this section, we list the results on sparse dataset SHHB. Tab. \ref{shhb} shows that we are the first place on F1-m (86.3$\%$).  Despite that there is a trace of backwardness on counting performance, we still derive a certain of improvement comparing with most related instance segmentation locator IIM. We dissert the crux to the baseline method. In segmentation localization, each predicted instance represents true semantic information, while the density map regressors cannot promise the responding value has the true semantic information. To be specific, a locator with high counting performance and low localization performance cannot be recognized as a good pedestrians learner. Moreover, there exists a contradiction phenomenon. With a higher localization precision, the more boxes proposals tend to be lost which incurs worse counting performance.
\begin{table}
\centering
\caption{Comparison with SOTA methods on sparse dataset.}
\renewcommand\arraystretch{1.3}
\setlength{\tabcolsep}{3mm}
\begin{tabular}{c|c|c|c} 
\whline
\multirow{2}{*}{Method} & \multirow{2}{*}{Backbone} & \multicolumn{2}{c}{SHHB}                                        \\ 
\cline{3-4}
                        &                           & \textbf{F1-m}/Pre./Rec. (\%)     & MAE/MSE                      \\ 
\whline
TinyFaces               & ResNet-101                & 71.1/64.7/79.0                   & -/-                          \\ 
\hline
RAZ Loc                 & VGG-16                    & 68.0/60.0/78.3                   & \underline{9.9}/\underline{15.6}                     \\ 
\hline
LSC CNN                 & VGG-16                    & 71.2/71.7/70.6                   & \textbf{8.1}/\textbf{12.7~}  \\ 
\hline
IIM                     & VGG-16                    & 80.2/84.9/76.0                   & 22.1/44.4                    \\ 
\hline
Ours                    & VGG-16                    & 83.8/89.4/78.0                   & 18.2/37.8                    \\ 
\whline
IIM                     & HR-Net                    & \underline{86.2}/\underline{90.7}/\underline{82.1}                   & 13.5/28.1                    \\ 
\hline
DCST                    & Swin-ViT                    & 86.0/88.8/\textbf{83.3~}         & 11.0/23.6                    \\ 
\hline
Ours                    & HR-Net                    & \textbf{86.3}/\textbf{91.9}/81.2 & 16.0/33.5                    \\
\whline
\end{tabular}
\label{shhb}
\end{table}

\subsection{Discussion}

\begin{figure*}[t]
    \centering
    \includegraphics[width=0.8\textwidth]{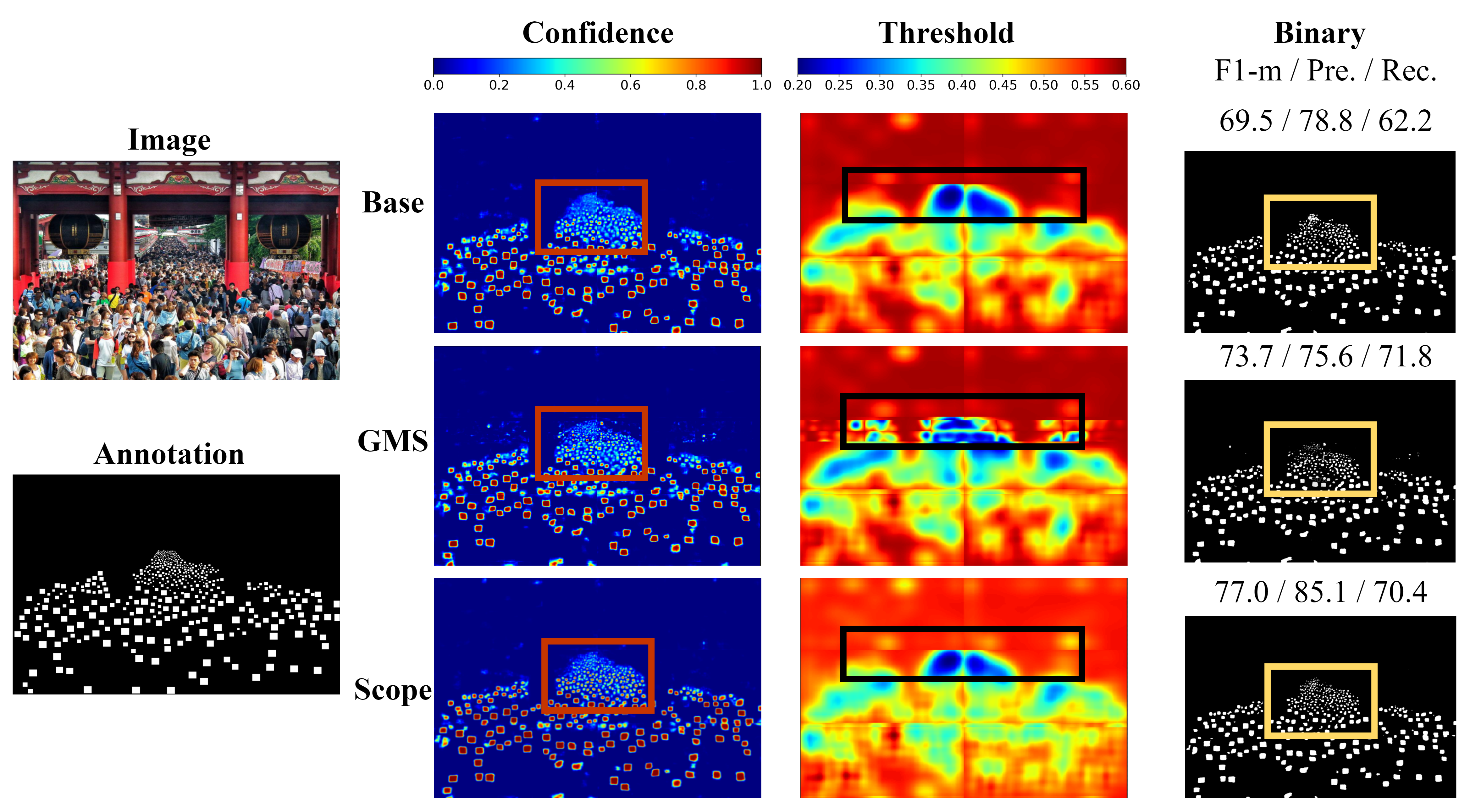}
    \caption{Visualization on typical sample from SHHA. The depicted information includes the confidence maps, threshold maps and binary maps predicted by baseline model (\textit{Base}), GMS inference (\textit{GMS}) and Scoped Teacher trained model (\textit{Scope}).}
    \label{fig_ctb}
\end{figure*}

In this section, we discuss how GMS and Scoped Teacher improves the final performance based on experiments in Section \ref{sec_exp}. {To facilitate clear discussion, we pick one typical sample from SHHA and visualize its predicted confidence maps, threshold maps and binary maps from models of baseline method, GMS inferred and Scoped Teacher transferred, as shown in Fig. \ref{fig_ctb}.}

To begin with Tab. \ref{tab AA}, we notice that directly implementing GMS in the testing time brings improvement. Thus, it demonstrates the effect of scale alignment. However, it goes as a common sense that the conventional image interpolation would not provide any additional information. {To this end, we argue that the improvement by GMS comes from distribution regularization and other forms of knowledge exploitation. See the column of Confidence in Fig. \ref{fig_ctb}, the red box selects a region filled with tiny instances. In the bottom of box, the GMS provides higher confidence than baseline. This is because the original representation of those instances with improved confidence is still in the latent space of crowd locator. Hence, this is the GMS exploited knowledge. Nevertheless, since there is only an effective resolution (dataset author provided resolution) of 1024 * 768 to the image in Fig. \ref{fig_ctb}, the instances in the top of the red box have representations of outliers, which are still outliers after alignment via GMS. With confidence variance by GMS explained, we put concentration on threshold learning. See the black box in the Threshold column of Fig. \ref{fig_ctb}, GMS brings unsmooth distribution to threshold map. Actually, the regularization of GMS does not introduce any influence on model parameters. In the right and left side of the black box, the two regions with obviously low thresholds should be negative. It is blamed on the poor robustness of the baseline model. Since the corresponding area in red box shows better prediction on confidence, the abnormal low threshold area in black box is hard to be explained by non-semantic distortion. In the Scoped Teacher training, GMS exploits the wrong prediction actively to teacher model to correct them, see black box in the Scope row of Fig. \ref{fig_ctb}. Thus, this is also the GMS exploited knowledge. }

Then, we discuss the effect of the Scoped Teacher. {There may be interests on how the knowledge transfers and what role of the Scoped Teacher plays beyond bridge in transform. Similarly, beginning from confidence analysis namely Confidence column in Fig. \ref{fig_ctb}, the red boxes select our ROI. We assert that Scoped Teacher guides student to build a connection between normal tiny representations with GMS aligned representations. This is because the student locator fed with original outliers is inclined to make prediction being similar with teacher fed with mapped embeddings and the process makes implicit mapping transform from outliers embedding to normal embedding. What’ s more, there is another interesting phenomenon that the confidences are higher in red boxes when comparing Scope column with GMS column. We argue that the Scoped Teacher makes a further aggregation to the knowledge. The scope of GMS is limited within the temporary input. However, the Scoped Teacher learns to build the connection from all similar representation in the training set. Thus, the knowledge in confidence is the connection and the role Scoped Teacher plays is the connector and aggregator. Then, we analyze the Threshold learning. See Threshold column in Fig. \ref{fig_ctb}, the black boxes also select our ROI. We notice that the threshold from Scope is more smooth than GMS. In consistency regularization, the negative regions outputs FP samples which exposes the un-robustness of locator and the corresponding loss is deployed to optimize the un-robustness. What’ s more, since GMS treats distribution discretely, it makes images lost physical features. The Scoped Teacher aids locator learn GMS exploited useful information and ignore the issues incurred by losing physical feature. Thus, the knowledge transferred is the punishment on un-robustness and the role Scoped Teacher plays is a filter to select useful knowledge.} Finally, we put analysis on Binary column of Fig. \ref{fig_ctb}. See the yellow boxes, the Scope depicts less boxes than GMS. And the recall comparison shows there are more boxes are removed, while the precision comparison shows there are more accurate boxes predicted. Thus, the knowledge transferred is box refinement and the role Scoped Teacher plays is the refiner.

\section{Conclusions}

This paper aims to tackle the essential issue, intrinsic scale shift in crowd localization. Specifically, we propose to regularize the chaotic scale distribution to align scale shift. Gaussian Mixture Scope (GMS) is proposed to implement the scale distribution regularization which is from distribution decoupling and alignment among sub-distributions perspective. Moreover, the GMS introduces spatial feature in regularization facilitating to geometrize the alignment which can thus be deployed via image interpolation. To further address the semantic distortion incurred by adaptative decoupling, we propose a novel sub-distribution re-aggregation strategy. What’ s more, a Scoped Teacher model with corresponding consistency regularization is further introduced to transfer knowledge from GMS processed data to locator and be a novel manner to implement GMS to make locator actively learn the knowledge. The proposed GMS is remarkably visible in improving localization performance. The Scoped Teacher model bridges between data with model to aid the implementation of GMS in training phase and promote final localization results. Extensive experiments show that the proposed work achieves state-of-the-art on popular datasets of the crowd localization. In the future, we will discuss how to align the average scale shift among datasets namely extrinsic scale shift, which is to locate the crowds towards the open-set.

\bibliographystyle{IEEEtran}

\bibliography{refer.bib}

%

\end{document}